\documentclass[conference]{IEEEtran}
\IEEEoverridecommandlockouts
\usepackage{cite}
\usepackage{amsmath,amssymb,amsfonts}
\usepackage{algorithmic}
\usepackage{graphicx}
\usepackage{float}
\usepackage{textcomp}
\usepackage{xcolor}
\usepackage{subfigure}
\usepackage{url}
\usepackage{hyperref}
\usepackage{algorithm}
\usepackage{algorithmic}
\usepackage{makecell}

\usepackage{cite}
\usepackage{amsmath,amssymb,amsfonts}
\usepackage{algorithmic}
\usepackage{algorithm}
\usepackage{graphicx}
\usepackage{textcomp}
\usepackage{xcolor}
\def\BibTeX{{\rm B\kern-.05em{\sc i\kern-.025em b}\kern-.08em
    T\kern-.1667em\lower.7ex\hbox{E}\kern-.125emX}}
\begin{document}

\title{Unified, User and Task (UUT) Centered Artificial Intelligence for Metaverse Edge Computing}


\author{\IEEEauthorblockN{Terence Jie Chua}
\IEEEauthorblockA{Graduate College\\Nanyang Technological University\\
terencej001@e.ntu.edu.sg }
\and

\IEEEauthorblockN{Wenhan Yu}
\IEEEauthorblockA{Graduate College\\Nanyang Technological University\\
wenhan002@e.ntu.edu.sg }
\and

\IEEEauthorblockN{Jun Zhao}
\IEEEauthorblockA{School of Computer Science \& Engineering\\ Nanyang Technological University\\
junzhao@ntu.edu.sg }
}

\maketitle

\begin{abstract}
The Metaverse can be considered the extension of the present-day web, which integrates the physical and virtual worlds, delivering hyper-realistic user experiences. The inception of the Metaverse brings forth many ecosystem services such as content creation, social entertainment, in-world value transfer, intelligent traffic, healthcare. These services are compute-intensive and require computation offloading onto a Metaverse edge computing server (MECS). Existing Metaverse edge computing approaches do not efficiently and effectively handle resource allocation to ensure a fluid, seamless and hyper-realistic Metaverse experience required for Metaverse ecosystem services. Therefore, we introduce a new Metaverse-compatible, Unified, User and Task (UUT) centered artificial intelligence (AI)-based mobile edge computing (MEC) paradigm, which serves as a concept upon which future AI control algorithms could be built to develop a more user and task-focused MEC.

\end{abstract}

\section{Introduction}
The term \textit{Metaverse} has a futuristic notion: one written in fictitious novels that feature a cyberpunk world. However, the Metaverse is not just a construed vision of the future but is fast becoming a reality. One can think of the Metaverse as an extension to the current web we use, in which the gap between the physical and virtual world is bridged by extended reality (XR) technologies. Equipped with XR-enabled wearables in the form of, but not limited to headsets, glasses, watches and gloves, Metaverse users can project their physical being, surrounding objects and motions onto the virtual world through digital twinning. The ability to accurately project our physical self, objects and actions onto the Metaverse is significant. Firstly, the ability to express oneself accurately improves the Metaverse socialization experience. Geographical distances do not limit interactions from all over the world; people can meet, connect and interact freely on the Metaverse in a manner close to an in-person experience. Secondly, and even more importantly, an accurate real-time digital twinning of real-world matter, actions and events brings forth possibilities for developing ecosystem services within the Metaverse.
\\

\subsection{XR experience within the Metaverse}
XR is the overarching term which encompasses virtual reality (VR), augmented reality (AR) and mixed reality (MR). These terms can be organized and ordered along a reality-virtuality continuum, where VR reflects the closest to an entirely virtual experience, followed by AR, then MR. VR platforms create a separate space within the virtual world in which VR users are fully immersed in the digital realm. On the other hand, AR platforms project virtual world objects onto physical world scenes as an overlay through the lenses of an XR-enabled wearable device. MR can be considered a sophistication of AR, in which users are not bounded by screens and can interact with virtual world objects, which coexist alongside physical world objects.


\subsection{Metaverse ecosystem services}
XR technologies empower the bidirectional relationship between physical world objects, people and motions and their virtual counterparts. This relationship is the driving force behind the establishment of Metaverse ecosystem services. Metaverse ecosystem services can be classified as (i) Physical Twin (PT)-driven and (ii) Digital Twin (DT)-driven services.

\subsubsection{PT-driven services} In PT-driven services, physical twin impacts and influences the digital twin. Examples of PT-driven services include Metaverse socialization applications where users connect in the Metaverse, through the mask of their avatars or holographic displays, in a hyper-interactive manner. The PT-DT relationship opens doors to interactions of more complex purposes such as virtual meetings, education, sporting events, alternative forms of entertainment, and teleconsultation.

\subsubsection{DT-driven services}In DT-driven services, digital twins are utilized or acted upon to effect decisions and changes on their physical counterparts~\cite{ahmadi2021networked}. For instance, real-time digital twinning allows the mapping of real-world traffic onto a virtual Metaverse map. Real-time digital twinning permits real-time, complex traffic and navigational planning through training artificial intelligence (AI) models on the historical data of physical twins. With a unified Metaverse map, one can devise advanced urban planning and design intelligent buildings. In manufacturing facilities, digital twins can be virtually controlled, either manually or via artificial intelligence, to effect changes to the physical twin (PT). It is beneficial in manufacturing or logistics, where a centralized view of the manufacturing process can improve workflow efficiency and output. Furthermore, pilot tests can be conducted before the launches of products and services through simulations with digital twins.

Lim \emph{et al.}~\cite{lim2022realizing} discuss PT-driven and DT-driven services in the Metaverse, supported by edge intelligence. DT-driven virtual city is further presented as a case study.

\subsection{Existing limitations}
\subsubsection{Computation-intensive applications}
The projection of the physical world onto the virtual world through XR and complex interactions between PT, their DT and other virtual beings or objects are computationally intensive. The rendering and control of DT through the PT, and vice-versa are not suitable for current user devices, as these devices lack the computational capability to power Metaverse applications. 

\subsubsection{Edge computing and its challenges}Mobile edge computing (MEC)  has been established as one of the most promising pillars to power the Metaverse and its applications~\cite{cheng2022will}. In a typical Metaverse application computation offloading scenario, images, audio, or other raw sensory information are collected by the Internet of Things (IoT) and sensors. In their raw or altered forms, this information is then transmitted to a Metaverse edge computing server (MECS) for replication and rendering of DTs and their interactions with the virtual world. These decisions and changes to the DT are then transmitted back to the user devices for PT response in the physical world. However, current edge computing technologies must be optimized as Metaverse applications are computationally demanding.

\subsubsection{Existing AI approaches and their limitations}Across different MEC stages, traditional convex optimization, game-theoretic methods and meta-heuristics have been adopted to establish a reliable and resource-efficient MEC. While computationally effective at handling smaller-scale systems, these approaches are less capable in handling much more complex, dynamic, and time-varying systems such as those encountered within Metaverse applications. Other works have adopted neural networks to solve MEC resource allocation problems. Deep reinforcement learning (DRL) approaches are commonly utilized in these works due to their ability to handle complex, sequential decision-making to accommodate the highly dynamic MEC systems~\cite{kibria2018big}. However, existing works mainly utilize basic DRL architectures to tackle smaller-scale, simple MEC resource allocation problems. These elementary architectures are not designed to handle Metaverse user or task-specific requirements. The Metaverse features varying applications which have different computation and communication resource requirements. Moreover, user devices have heterogeneous hardware specifications and individuals have varying preference for Quality of Experience (QoE) factors. It is pertinent to design a user and task-centered AI-based MEC for Metaverse applications.

\begin{figure*}[t]
\centerline{\includegraphics[width=0.8\linewidth]{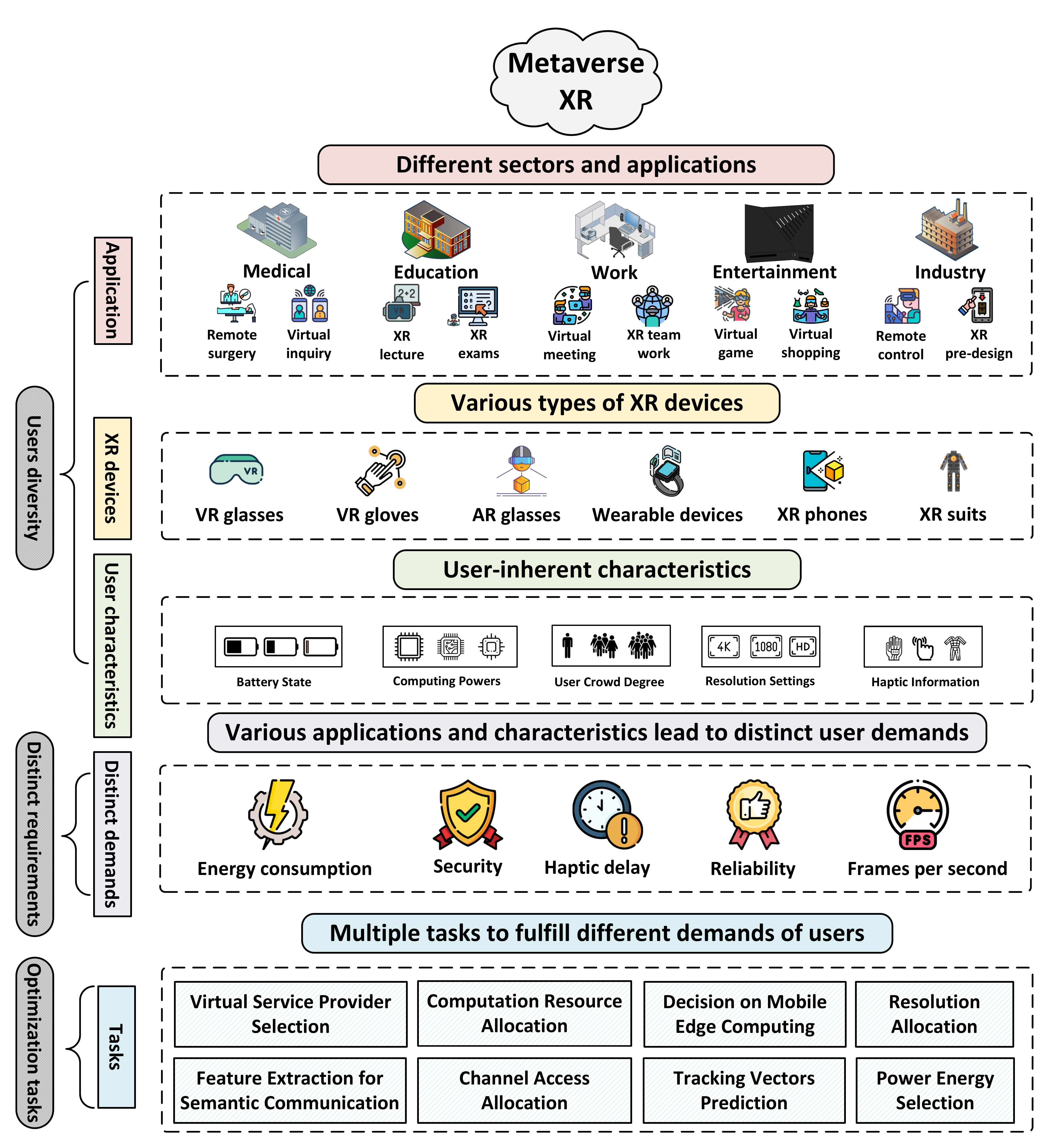}}\vspace{-5pt}
\caption{Diagram showing the driving Metaverse user diversity, their influence on resource demands, and the MEC tasks to fulfill them.}
\label{fig:systemmodel}
\vspace{-0.3cm}
\end{figure*}

\section{Metaverse Quality of Service (QoS) Requirements}
Metaverse and its applications have to fulfil the QoS requirements in order for it to be ubiquitous. These requirements are characterized in both qualitative and quantitative manner.

\subsection{Qualitative Requirements}
\subsubsection{Bridge between Real and Virtual World}
One key feature of the Metaverse which separates it from the current web is the trans-reality experience and interconnectivity between the physical and virtual world. This integration permits the development and function of Metaverse ecosystem services.

\subsubsection{Unified world}
The barriers between distinct virtual worlds have long impeded cross-world interactions, transactions and collaborative developments. The existence of a unified world, such as the Metaverse, opens opportunities for developing extensive Metaverse ecosystem services.

\subsubsection{Immersive Experience}
Risk-sensitive Metaverse applications such as intelligent traffic and telesurgery require a highly fluid, high-resolution, seamless experience.

\subsection{Quantitative Requirements}
Metaverse applications have stringent technical requirements to deliver a seamless, fluid, high-quality and fully immersive experience to the masses~\cite{cheng2022will,xu2022full}.

\subsubsection{Visual display}Metaverse applications are required to deliver high-resolution scenes at 64 M pixels, with a horizontal and vertical fields of vision at 150\textdegree~and 120\textdegree, respectively. These wide-angle, high-resolution scenes must be displayed at 120 frames per second to ensure a true-to-life experience. Such high-resolution, field-of-view, and frame-rate scenes demand a scene rendering rate of 1 Gbps.

\subsubsection{Motion-to-Photon (MTP)}Motion-to-Photo (MTP) is the time lag between a movement in the XR-enabled device and the visual display in the user's device~\cite{elbamby2018toward}. MTP latency of greater than 20ms would cause user discomfort~\cite{cheng2022we}.

\subsubsection{Tactile experience} Real-time transmission of true-to-life haptic feedback (less than 1 ms delay) to Metaverse application users is critical to allow users to have a true-to-world experience~\cite{hou2021intelligent}. Furthermore, haptic feedback transmission requires a relatively low packet error rate.

To ensure that Metaverse users have a high quality, true-to-life visual experience and sensation, MEC empowering Metaverse applications have to ensure ultra-high reliability and ultra-low latency in communication and computation~\cite{cheng2022we, lee2021all}. These features can be enforced through a unified, user and task-centered AI-based-MEC for the Metaverse. Interested readers can refer to~\cite{yu20226g,cheng2022will} for more discussions of Metaverse QoS requirements.

\begin{figure*}[t]
\centerline{\includegraphics[width=1\linewidth]{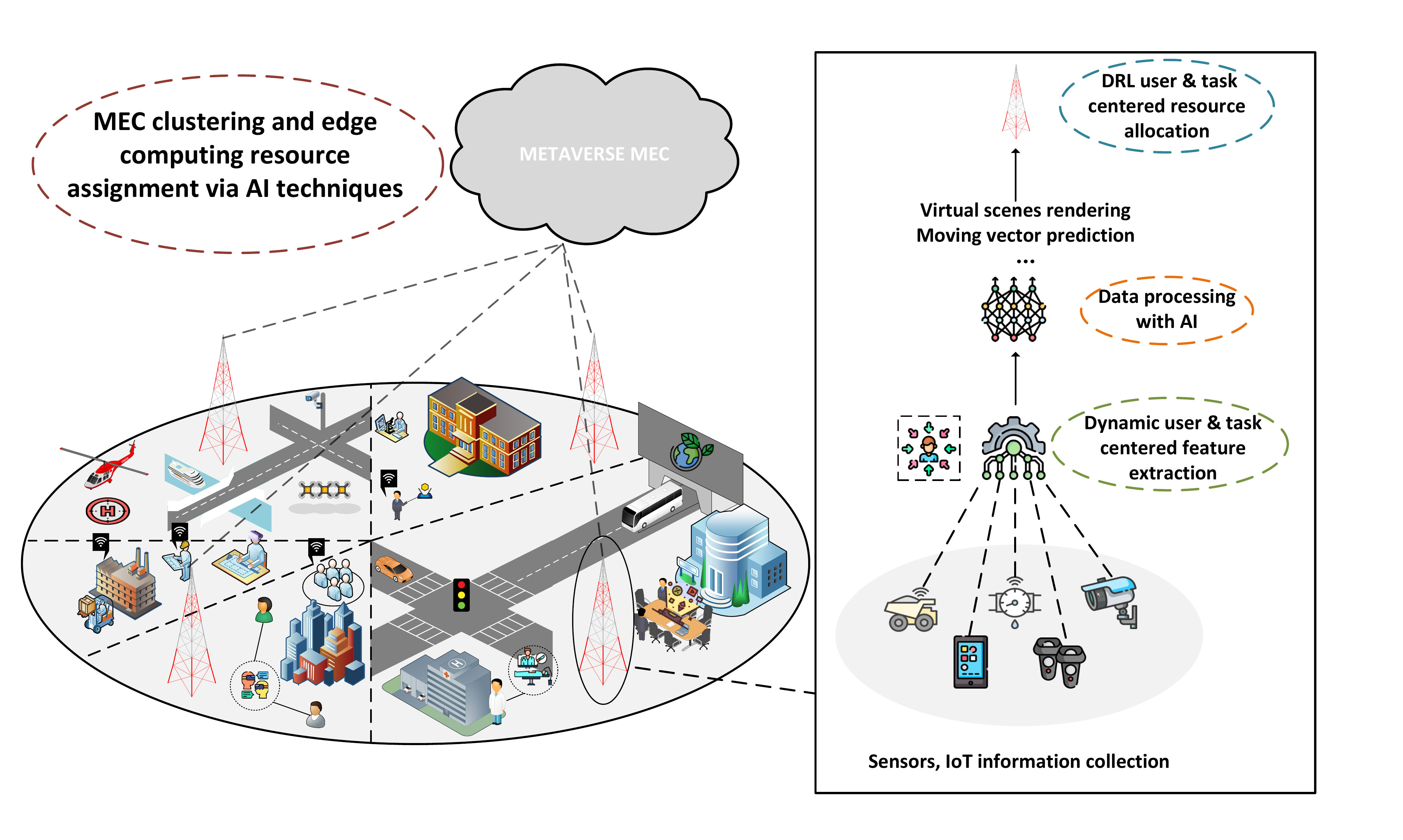}}\vspace{-5pt}
\caption{Unified, User and Task (UUT)-centered MEC system model.}
\label{fig:MEC}
\vspace{-0.3cm}
\end{figure*}

\section{Unified, User \& Task Centered Artificial Intelligence}

When designing a user-centered AI-based MEC system for Metaverse applications, it is important to understand the underlying factors driving variation in user resource demands. Firstly, the Metaverse hosts a myriad of Metaverse applications, where the MEC computation and communication resource required to enable each application varies. Metaverse applications span across various sectors, including medical, education, work, entertainment and industrial (as shown in Fig.~\ref{fig:systemmodel}). Naturally, the sector-specific applications have different levels of computation and communication demands. For instance, remote surgeries require extremely high visual resolution and frame rate, alongside with hyper-realistic and ultra-low latency tactile feedback as it is a safety-sensitive application. On the other hand, an XR lecture is not safety-sensitive and there is higher tolerance for computation and communication delays.

In addition, there are variations across XR device hardware specifications. User devices such as VR glasses, gloves, XR phones and suits have different computation power, communication capabilities and battery capacity. These variations are to be taken into account when designing user and task-centered optimization algorithms for MEC of Metaverse applications.

User preferences and characteristics are to be considered when designing a user and task-centered optimization algorithm for MEC. Priorities for lower battery consumption, higher scene resolution, and realistic haptic sensations vary across users. Furthermore, the mobility patterns of users have to be considered to account for crowding of Metaverse users and minimizing user-MECS handovers.

The computation demands of the Metaverse application, limitations of XR devices and user characteristics drive variations in computation and communication demands between users. Some of the  demands include minimizing device energy consumption, haptic delays, and improving frame-rate, security and reliability of the Metaverse experience.

In order to meet the computation and communication demands of users, the MEC process for Metaverse applications must be optimized. Some aspects which can be optimized include the decision on whether a user's task is to be offloaded, MECS selection, and channel access allocation. Other factors (tasks) which have to be optimized include the selection of displayed scene resolution, selection of power output for computation offloading, feature extraction for semantic communication and tracking vector predictions.

To enable a Unified, User and Task (UUT)-centered AI-based Mobile edge computing (MEC) for Metaverse applications, several aspects of MEC must be optimized to provide users with a personalized, seamless, hyper-realistic experience. User and task-centered artificial intelligence can be integrated to optimize the different stages within the Metaverse MEC process~\cite{cao2019intelligent} as shown in Fig.~\ref{fig:MEC} and are elaborated in subsequent subsections. The different stages proposed in our article are inspired by the works of Yang \emph{et al.}~\cite{yang2020artificial} and Ren \emph{et al.}~\cite{ren2019edge}.

\subsection{Sensors, devices, IoT}
Physical devices such as XR-enabled wearables serve as sensors that capture sensory information in the form of audio, visual, touch, and kinesthetic. This information is generally size-wise significant as it contains detailed information about  the surroundings and an object or user's position relative to its surroundings. This large data size impedes edge computing capabilities as the transmission of large data sizes incurs latency, compromising the fluidity of experience expected within the Metaverse. Nevertheless, in several applications, such as virtual shopping or XR lectures, the user experience and performance are less sensitive to the loss of sensory details. Instead, it is the semantic information which is transmitted that is of greater importance. Machine learning techniques such as principle component analysis (PCA) or isometric mapping (ISOMAP) have long been adopted for data compression and feature extraction. Such techniques are able to extract the most meaningful semantic information~\cite{luo2022semantic}. In recent years, neural networks in the form of autoencoders have been employed for data compression. Depending on user or Metaverse application requirements, individual users may have varying user device battery consumption, scene resolution and frame-rate requirements. As such, deep reinforcement learning (DRL) techniques can be applied to dynamically organize the extent of data compression following user and Metaverse application requirements, bringing forth a \textbf{user-centered AI-based} feature extraction and semantic information transmission.


\begin{figure*}[t]
\centering
\subfigtopskip=2pt
\subfigbottomskip=2pt

\subfigure[Structure of User-centered reinforcement learning.]{
\begin{minipage}[t]{0.5\linewidth}
\centering
\includegraphics[width=1\linewidth]{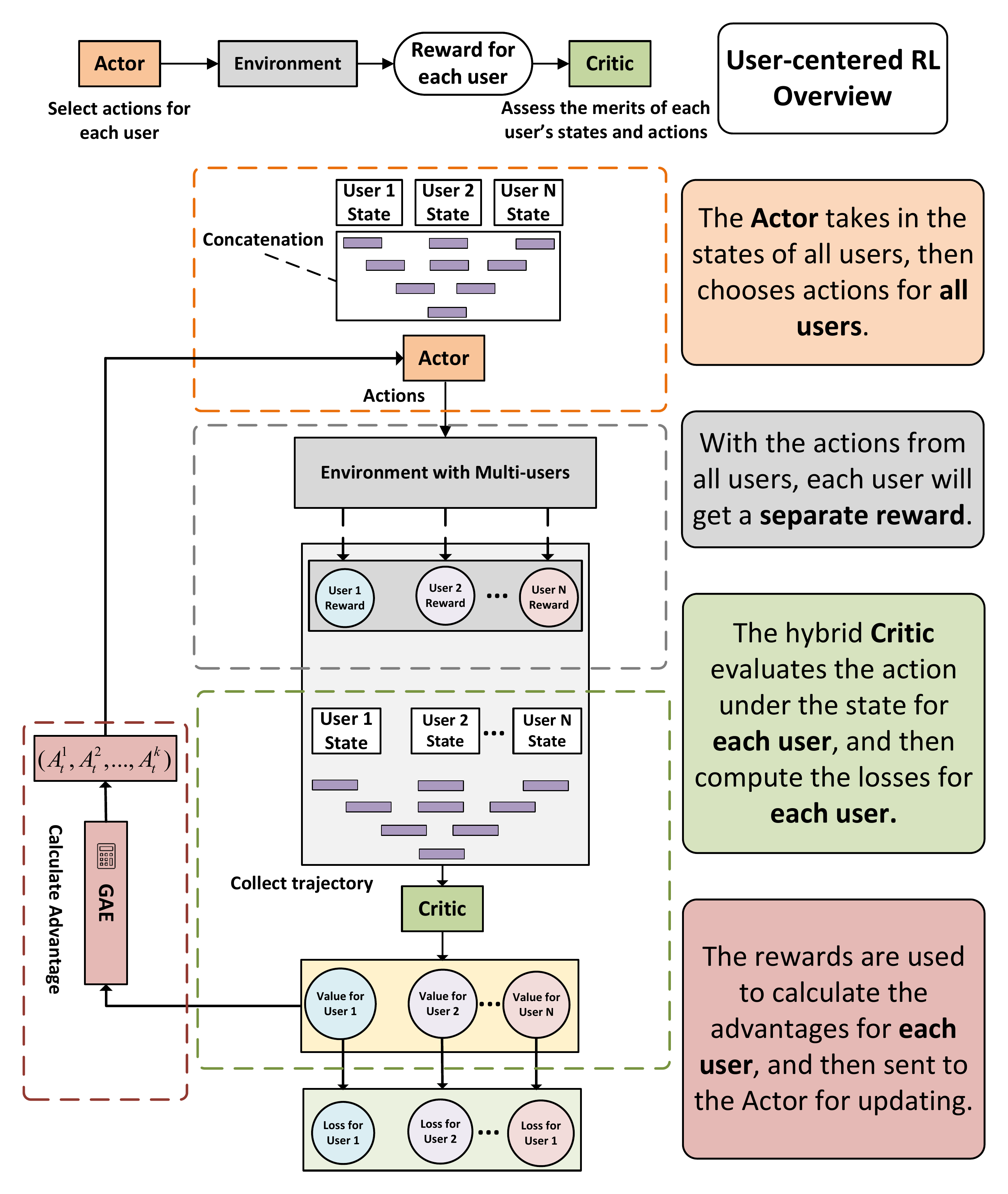}
\label{fig:UC}
\vspace{-10mm}
\end{minipage}
}%
\subfigure[Structure of Task-centered reinforcement learning.]{
\begin{minipage}[t]{0.5\linewidth}
\centering
\includegraphics[width=1\linewidth]{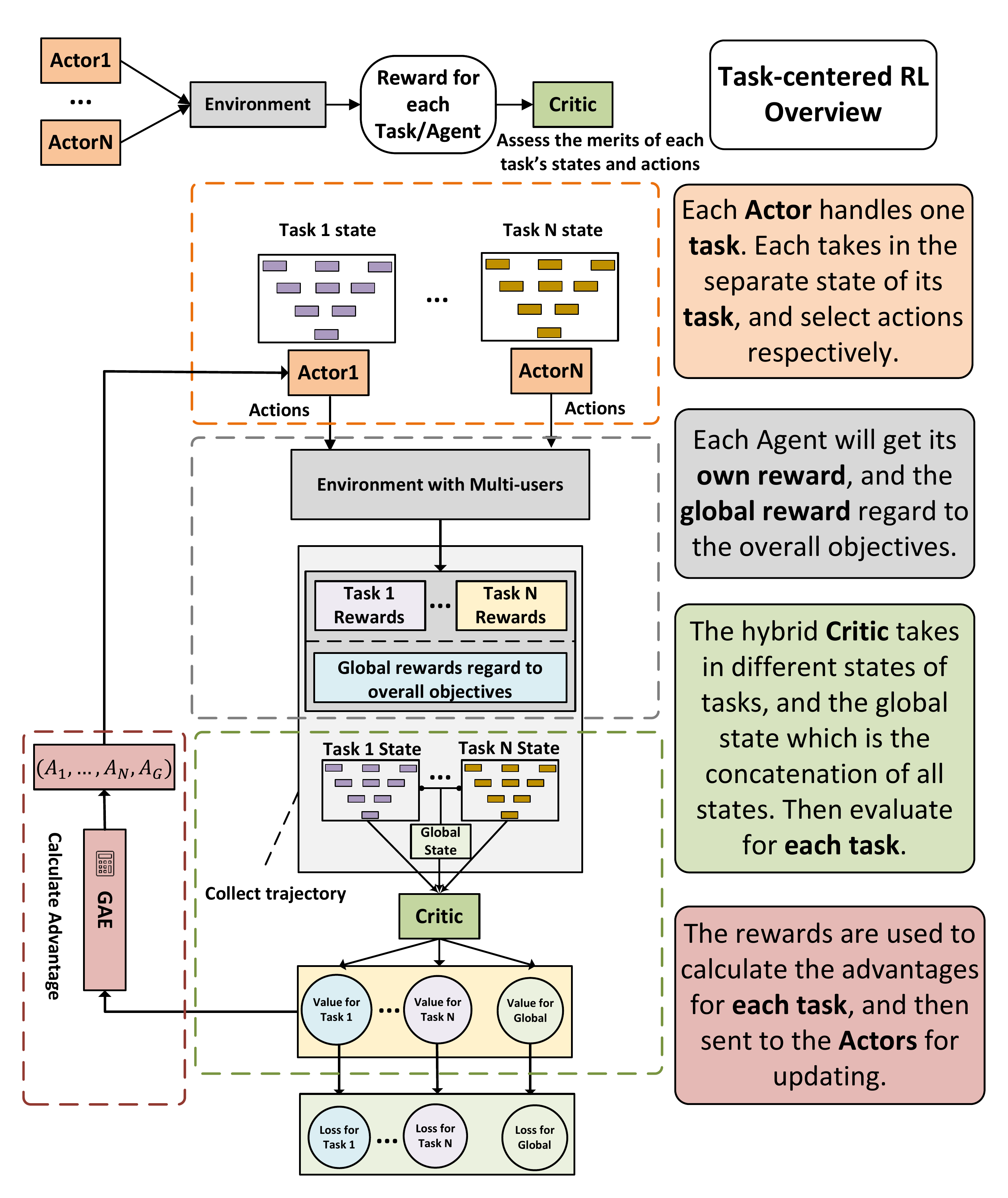}
\label{fig:TC}
\vspace{-10mm}
\end{minipage}
}%

\caption{User-centered DRL and Task-centered DRL.}
\label{fig:UCTC}
\vspace{-0.5cm}
\end{figure*}

\subsection{Metaverse Physics and Integration}
Metaverse applications are expected to be highly automated. The physics underlying the animation and the interaction between virtual world beings and objects must not be rule-based, as rule-based approaches to handling interactions between multiple entities within the Metaverse may not be sufficient. Such complex interactions rely on general artificial intelligence techniques such as deep reinforcement learning (DRL). A \textbf{user-centered} DRL agent, coupled with a segmentation model, can be implemented to predict user movements and behaviors (tracking vectors) based on historical tracking information to efficiently allocate resources to render scenes. An example is to use data-driven approaches to predict users eye movement patterns and place emphasis and resources on rendering scenes within the point of fixation, and less on scenes on the peripheral~\cite{patney2016towards}.



\subsection{MECS Smart Resource Allocation}
The MECS smart resource allocation layer handles the MEC resource allocation process for each application. Within a Metaverse application, multiple users will offload raw or altered sensory data to a Metaverse edge computing server (MECS) for computation. How does the MECS ensure that each user has a fluid, seamless, hyper-realistic sensory experience while ensuring that the user devices have sufficient battery charge to keep up with the continuous integration and interaction with the Metaverse? The MECSs' objectives are convoluted, and the users' behaviours are highly dynamic. Such scenarios have been tackled in several works which employ DRL techniques. DRL is considered state-of-the-art as it can handle complex, highly dynamic and constantly evolving sequential decision-making.

Existing DRL techniques applied to solving edge computing challenges adopt simple algorithms and architectures with little consideration of the needs of individual Metaverse users. In a prototype system where few users are involved, the consequences of adopting elementary neural network structures within DRL, are trivial. Extending these AI techniques to manage users in large-scale systems is unrealistic with rudimentary architectures. \textbf{Firstly}, for discrete actions, the number of possible action outcomes grows exponentially with the number of users. The sparsity of the action space renders a decision for actions highly computationally inefficient. \textbf{Secondly}, existing architectures do not account for heterogeneous requirements, objectives, and constraints among users. In addition, user devices may have different state input dimensions, which render existing architectures inadequate.



A \textbf{user-centered} AI-based MEC for Metaverse applications should take into consideration the computation demands of the Metaverse application, limitations of XR devices, and user characteristics and preferences. A user-centered edge computing smart resource allocation algorithm is built upon existing DRL algorithms such as Deep Q-Networks (DQN), Advantage Actor Critic (A2C), and state-of-the-art algorithms such as Proximal Policy Optimization (PPO). While Actor-Critic algorithms are structurally different from its preceding counterparts, the user-centered ideology is applicable. In contrast to existing DRL architectures where a single reward is obtained from the neural network, multiple tails, each associated with a user is designed to assess the value of each user's action, and assign a reward accordingly for each user (as shown in Fig.~\ref{fig:UC}).

In the Actor-Critic variants of a user-centered MEC, the architectural changes extend to the Critic. The most notable alteration is found in the output layer of the Critic, in which state values $V_i$ for each user are produced, in contrast to a single unified state value $V$ in traditional RL architectures. These user-centered state values are then used to compute user-centered loss $L_i$ and advantage values $A_i$ for update of the Critic and Actor, respectively. The detailed and minuscule observations of each user improve the overall model in handling individual user demands.

\subsection{Unification of Tasks for MEC of Metaverse Applications}
Apart from developing a user-centered AI-based MEC control algorithm and structure to handle a large-scale user-base Metaverse, one has to consider the synchronization of several decision-making agents. AI-based MEC for Metaverse applications demands the optimization of several computation and communication resources, and consequently involves multiple decision-makers, each handling different tasks (described in Fig.~\ref{fig:systemmodel}). As each decision making agent serves to optimize a factor with regards to an objective, it may be executed in a manner irrespective of other agents. These decision-makers for each task are often considered separate entities in existing works. In actuality, the decisions made by agents at the control layer are not independent, as they are influenced by the decisions made by other decision-making agents. There is a need to develop a \textbf{task-centered AI-based} resource allocation MEC system to ensure that the sub-objectives of underlying MEC tasks are fulfilled alongside a global objective.

In the computation offloading of Metaverse applications, there is often more than one decision-making agent. For instance, an MECS DRL orchestrator may incorporate several DRL decision makers, each controlling different variables of concern. In the \textbf{Task-centered} RL architecture, several Actors, each making decisions pertaining to a MEC-related task, will take in states and output task-specific actions. Note that the states for each Actor may be different. Each agent representing a task will be rewarded based on their sub-objectives. 

Similar to the User-centered DRL, the architectural changes extend to the Critic for the Actor-Critic variants (shown Fig.~\ref{fig:TC}). The highlight of the task-centered MEC AI architecture is the multi-head Critic. The Critic takes in the differing Actor states alongside a concatenation of the Actors' states and outputs rewards, advantage, and loss values for each Actor. A global reward is also issued to assess the unified merit of the agents' actions (merit of actions taken for each task).




\subsubsection{User-centered Case Study}
To verify the validity of the User-centered RL framework, we propose a User-centered MEC to tackle a VR scenes computation offloading task, where the goal is to minimize overall scene transmission latency via channel access allocation (task). We assign users with distinct requirements pertaining to frames-per-seconds (FPS), tolerable transmission delay, tolerable device energy consumption, and device computation capabilities. We assume that each user is able to render VR scenes on their device locally, or via an MECS. We employ a User-centered RL architecture (Fig.~\ref{fig:UC}), traditional RL architecture, and an agent which performs random actions to compare the performance of each algorithm. The comparison is bench-marked based on the VR scenes rendering success rate (\% of VR scenes successfully rendered per second). From the results (Fig.~\ref{fig:UCcase}), it is observed that the User-centered RL achieves a much higher task success rate when compared to the traditional RL algorithm, across the training process. This emphasizes the superiority of a User-centered RL in tackling challenges with heterogeneous user requirements.


\subsubsection{Task-centered Case Study}
\begin{figure*}[t]
\centering
\subfigtopskip=2pt
\subfigbottomskip=2pt

\subfigure[Performance of User-centered DRL.]{
\begin{minipage}[t]{0.45\linewidth}
\centering
\includegraphics[width=1\linewidth]{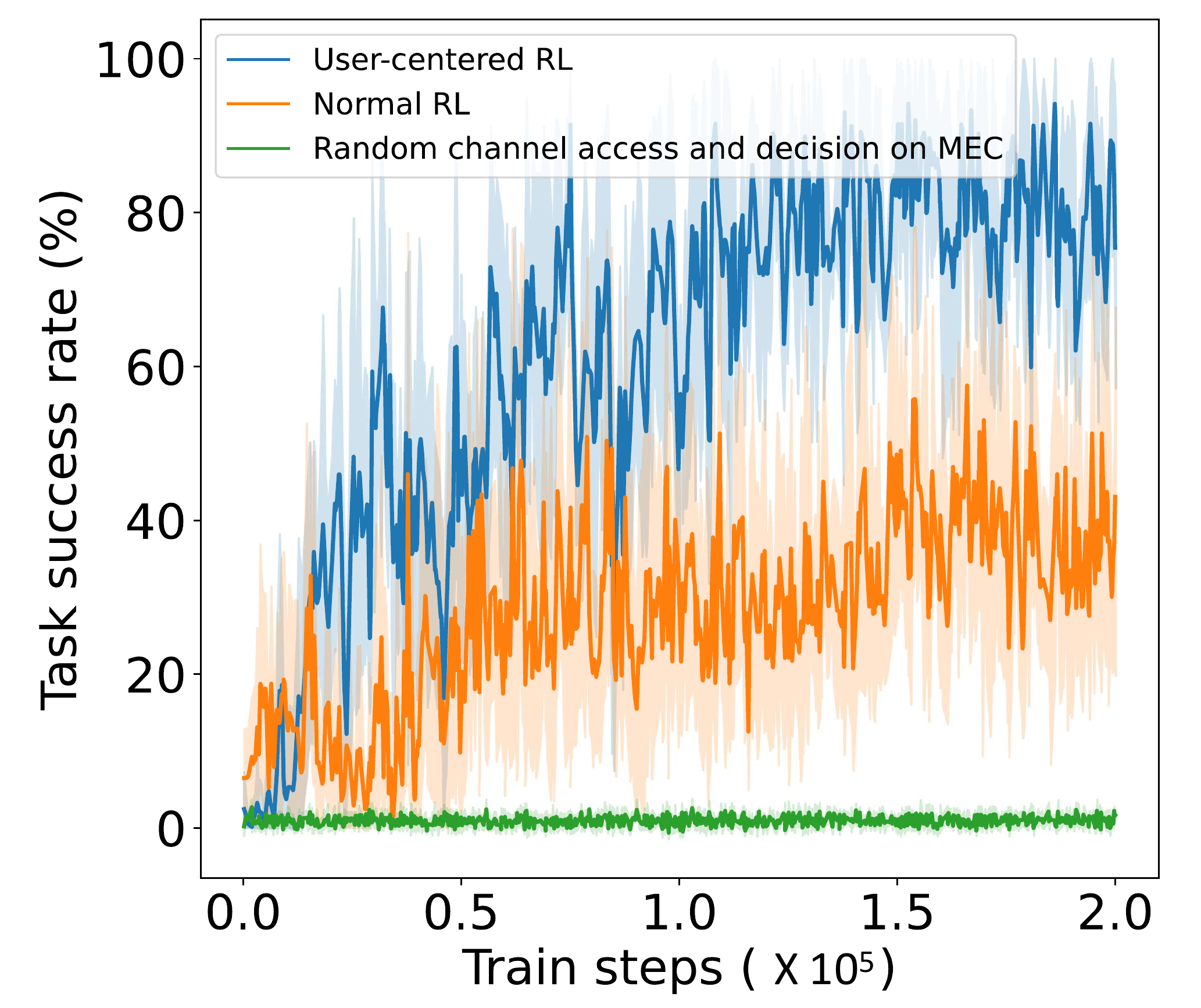}
\label{fig:UCcase}
\vspace{-10mm}
\end{minipage}
}%
\subfigure[Performance of Task-centered DRL.]{
\begin{minipage}[t]{0.45\linewidth}
\centering
\includegraphics[width=1\linewidth]{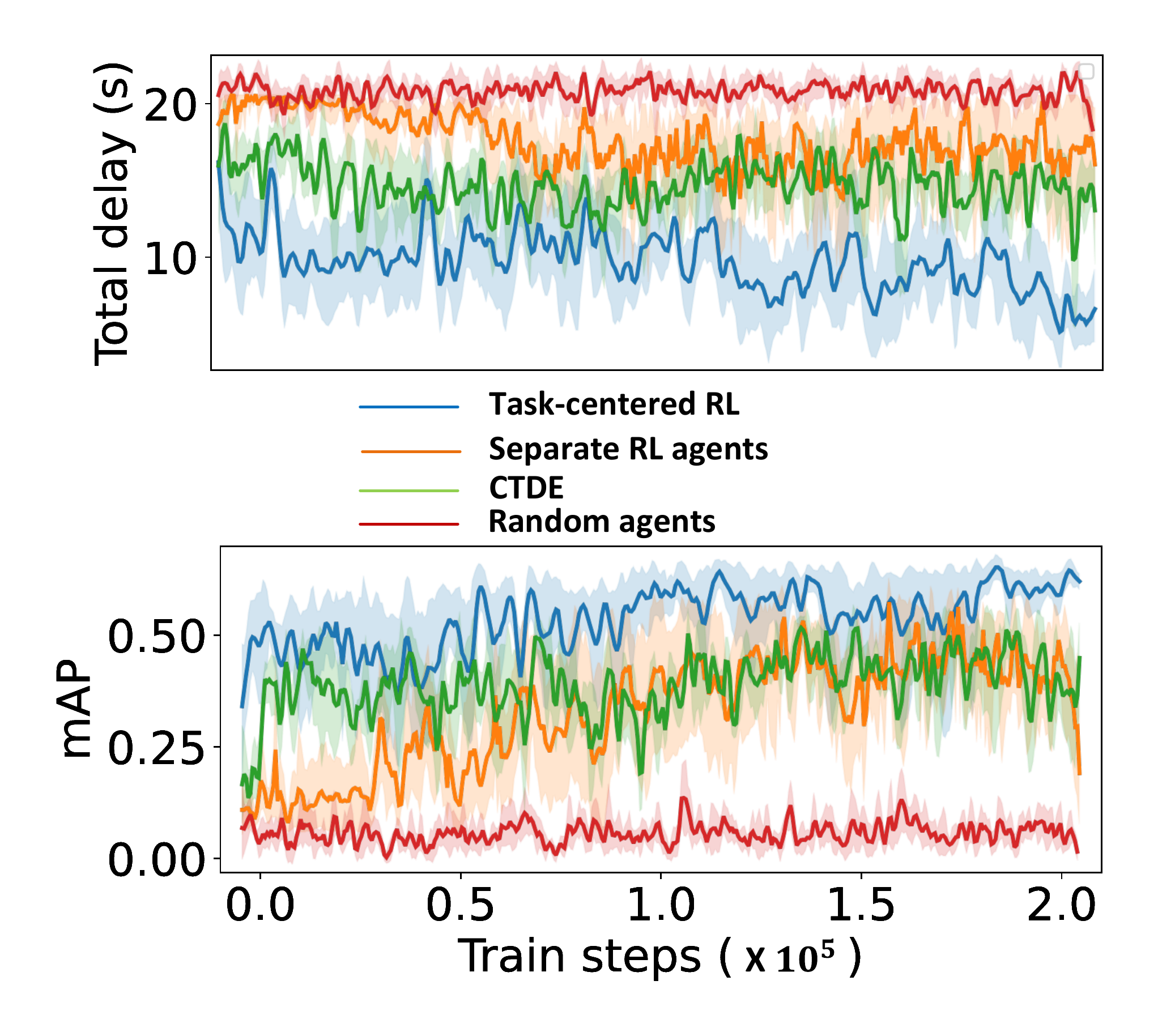}
\label{fig:TCcase}
\vspace{-10mm}
\end{minipage}
}%

\caption{Performance of User-centered DRL and Task-centered DRL.}

\vspace{-0.5cm}
\end{figure*}

To examine the feasibility of our Task-centered RL architecture, we proposed a Metaverse digital twinning scenario in which smart vehicles are travelling along the roads, detecting signages and traffic information to be uploaded to the MECS for the development of a Metaverse Map. The vehicles' objectives are to minimize scene uplink transmission latency while maximizing object detection accuracy. The MEC-related optimization variables (tasks) are MECS-user allocation and uplink transmission selection. From Fig.~\ref{fig:TCcase}, it is clear that the task-centered DRL algorithm achieves the lowest delay and highest mean Average Precision (mAP) score across the training process. This shows that the task-centered DRL algorithm manages to achieve the highest object detection score at the lowest transmission latency.

\subsection{UUT Smart Allocation}
Developing a Unified, User and Task (UUT)-centered AI-based MEC will empower Metaverse applications. Our proposed User-centered and Task-centered DRL architectures (Fig.~\ref{fig:UCTC}) can be combined together to build a UUT-centered DRL architecture. Similar to the User-centered DRL structure, individual user states are concatenated and fed into the Actor. The Actor for each task will then produce a reward for each user. Subsequently, the multi-head Critic will take in concatenated user-states across different tasks and output advantage and loss values for each user-task configuration (User $\times$ Task number of outputs) to update the UUT Actors and Critics.


\subsection{Centralization of Metaverse Applications}
A central cloud server is the primary controlling entity that manages several MECS in charge of different Metaverse applications. The central cloud server handles less dynamic and time-sensitive background computations that are persistent across different Metaverse applications~\cite{lim2022realizing}. An example of a computation job may be knowledge discovery and data mining, which seeks to uncover deeper patterns and trends in Metaverse application, task and user resource demands. Imbued with AI capabilities, the cloud server can achieve an overall more efficient resource allocation by seeking similarities between Metaverse applications through clustering methods and reallocating more MECS to handle resource-demanding applications.

\section{Future Directions}

We now identify a few directions for future research.

\subsection{Alternative Approaches}The vision of having a Unified, User and Task-centered AI-based MEC encompasses many other forms of alterations to and evolutions of existing DRL algorithms. The structural changes to traditional DRL algorithms presented in this article are non-trivial. Further investigations into UUT-centered DRL architectures or algorithms would propel this field forward.

\subsection{Redefining User-perceived QoE}The Metaverse is yet to be widely adopted, and the feedback on user experience is not yet well surveyed. Future works should study users' expectations of the virtual world and its applications. A thorough study into user requirements and preferences can assist Metaverse Mobile edge computing service providers in resource planning to cater to users' needs and improve the QoE.

\subsection{Privacy}The development of user-centered AI-based MEC raises concerns about leakages of individuals' private information. Potential leakages cause concern as user-tailored DRL involves neural network layers adjusted to cater to individuals' needs. Subsequent works could consider alternative forms of privacy-preserving edge computation, such as dividing the decision-making agents' neural networks into two components, sets of local layers and a centralized basal network. This disconnect could prevent potential leakages of user characteristics.



\section{Conclusion}
This article provides an overview of existing and potential Metaverse applications and the communication and computation requirements to enable them. We then introduce a Unified, User and Task (UUT)-centered AI-based MEC paradigm for Metaverse applications, which would form the bedrock for future developments within a complex and dynamic, multi-user and multi-task Metaverse edge computing.

{\small
\bibliographystyle{IEEEtran}
\bibliography{ref}
}

\end{document}